\documentclass[runningheads]{llncs}
\usepackage{graphicx}
\usepackage{csquotes}

\begin{document}
\title{Deep Learning based Frameworks for Handling Imbalance in DGA, Email, and URL Data Analysis}
\titlerunning{Cost-Sensitive Deep learning based Framework for Cyber Security}
%
%

\author{Simran K\inst{1}\and Prathiksha Balakrishna\inst{2} \and
Vinayakumar Ravi\inst{3,1}\and
Soman KP\inst{1}}
\authorrunning{Simran et al.}
%
\institute{Center for Computational Engineering and Networking, Amrita School Of Engineering, Amrita
vishwa vidyapeetham, Coimbatore, India. \email{simiketha19@gmail.com} \and Graduate School, Computer Science Department\\ Texas State University. \email{prathi.93april8@gmail.com}\and Center for Artificial Intelligence, Prince Mohammad Bin Fahd University, Khobar, Saudi Arabia.\\ \email{vinayakumarr77@gmail.com}}

\maketitle              
%
\begin{abstract}
Deep learning is a state of the art method for a lot of applications. The main issue is that most of the real-time data is highly imbalanced in nature. In order to avoid bias in training, cost-sensitive approach can be used. In this paper, we propose cost-sensitive deep learning based frameworks and the performance of the frameworks is evaluated on three different Cyber Security use cases which are Domain Generation Algorithm (DGA), Electronic mail (Email), and Uniform Resource Locator (URL). Various experiments were performed using cost-insensitive as well as cost-sensitive methods and parameters for both of these methods are set based on hyperparameter tuning. In all experiments, the cost-sensitive deep learning methods performed better than the cost-insensitive approaches. This is mainly due to the reason that cost-sensitive approach gives importance to the classes which have a very less number of samples during training and this helps to learn all the classes in a more efficient manner.

\end{abstract}
\section{Introduction}

Cyber Security is an area which deals with techniques related to protecting data, programs, devices, and networks from any attack, damage, and unauthorized access \cite{22}. There are various methods in  Cyber Security to secure systems as well as networks. The classical method is a signature-based system. A signature-based system relies on regular expressions which give domain-level knowledge. The main issue is that these signature-based systems can identify only described malware and cannot detect novel types of attacks and even existing variant types of attacks. In order to detect \enquote{0-day} malware, the researchers have followed the application of machine learning \cite{23}. 

Recently, the application of deep learning architectures are employed in  Cyber Security use cases and these models can extract features implicitly whereas machine learning algorithms require manual feature engineering \cite{24}, \cite{25}, \cite{26}, \cite{27}. To classify DGA generated domains, different deep learning approaches were proposed in \cite{21}, \cite{22}, and \cite{23}. The real-time datasets are highly imbalanced in nature and in order to handle it properly the concept of data mining approaches can be used. In this direction, numerous amount of research is being performed. In \cite{2}, an LSTM based model is proposed to handle multiclass imbalanced data for detection of DGA botnet. In this work, the LSTM model is adapted to be cost-sensitive and performed better than other cost-insensitive approaches. The main objectives of this paper are:

\begin{enumerate}
   \item This work proposes a cost-sensitive deep learning based framework for Cyber Security.
   \item  The performance of cost-sensitive approaches are evaluated on three different use cases in security namely, DGA, Email, and URL.
    \item The performance of cost-sensitive models are compared with cost-insensitive models.
    \item Various hyperparameter tuning methods are employed to identify the optimal network parameters and network structures.
    
\end{enumerate}
 
The remaining of this paper is arranged in the following order:
Section 2 documents a survey of the literature related to DGA, URL, and Email followed by background related to NLP, deep learning, and cost-sensitive concepts in Section 3. Section 4 provides a description of dataset. Section 5 describes the details of the proposed architecture. Section 6 reports the experiments, results, and observations made by the proposed architecture. Section 7 concludes the paper with remakes on future work of research.

\section{Literature Survey on deep learning based DGA, URL, and Email data analysis}
\subsection{Domain Generation Algorithms (DGAs)}

In \cite{1}, an LSTM network was proposed for real-time prediction of DGA generated domains. This work implemented binary as well as multiclass classification of DGA. This network has a detection rate of  90\% and a false positive (FP) rate of 1:10000.  In \cite{3}, convolutional neural network (CNN) and LSTM deep learning models were utilized to classify large amounts of real traffic. Simple steps were followed to obtain pure DGA and non-DGA samples from real DNS traffic and achieved a false positive rate of 0.01\%. In \cite{4}, deep learning based approach was proposed to classify domain name as malicious or benign. Performance of various deep learning techniques like recurrent neural network (RNN), LSTM, and classical machine learning approaches were compared. A highly scalable framework was proposed by \cite{5} for situational awareness of  Cyber Security threats. This framework analyses domain name system event data. Deep learning approaches for detection and classification of pseudo-random domain names were proposed in \cite{6}. Comparison between different deep learning approaches like LSTM, RNN, I-RNN, CNN, and CNN-LSTM was performed. RNN and CNN-LSTM performed significantly better than other models and got a detection rate of 0.99 and 0.98 respectively. A data-driven approach was utilized in \cite{7} to detect malware-generated domain names. This approach uses RNN and has achieved an F1-score of 0.971. A combined binary and multiclass classification model was proposed in \cite{2} to detect DGA botnet. This model uses LSTM network and has the capability to handle imbalanced multiclass data. A comprehensive survey on detection of malicious domain using DNS data was performed by \cite{8}.

\subsection{Uniform Resource Locator (URL)}

A comprehensive and systematic survey to detect malicious URL using machine learning methodologies was conducted by \cite{9}. URLNet was proposed in \cite{10} which is an end-to-end system to detect malicious URL. This deep learning framework contains word CNNs as well as character CNNs and has the capability to learn nonlinear URL embedding directly from the URL. In \cite{11}, various deep learning frameworks such as RNN, I-RNN, LSTM, CNN, CNN-LSTM were utilized to classify real URL’s into malicious and benign at the character level. LSTM and CNN-LSTM performed significantly better than other models and achieved an accuracy of 0.9996 and 0.9995 respectively. A comparative study using shallow and deep networks was performed by \cite{12} for malicious URL’s detection. In this work, CNN-LSTM network outperformed other networks by achieving an accuracy of 98\%. In \cite{13}, three models namely, support vector machine (SVM) algorithm based on term frequency - inverse document frequency (TF-IDF), logistic regression algorithm and CNN  based on the word2vec features were used to detect and predict malicious URLs. An online deep learning framework was proposed in \cite{14} for detecting malicious DNS and URL. The framework utilized character-level word embedding and CNN. In this work, a real-world data set was utilized and the models performed better than state-of-art baseline methods. URLDeep was proposed in \cite{15} to detect malicious URL’s. This deep learning framework based on dynamic CNN can learn a non-linear URL address. A cost-sensitive framework firstfilter was proposed in \cite{16} to detect malicious URL. This network can handle large-scale imbalanced network data.

\subsection{Electronic mail (Email)}

A complete review for filtering of email spam was proposed in \cite{17}. Machine learning based techniques and trends for email spam filtering were also discussed in this work. In \cite{18}, a machine learning based approach was proposed to classify email space. This framework basically classifies an email into spam and non-spam. This work also proposed a platform-independent progressive web app (PWA). In \cite{19}, deep learning based frameworks were proposed for email classification. This work proposed LSTMs and CNNs network which outperformed baseline architectures. CNN performed better than LSTM and achieved an F1 score of 84.0\%. \cite{20} proposed a multi-modal framework based on model fusion (MMA-MF) to classify email. This model fuses CNN and LSTM model. Image part of the email is processed by the CNN model whereas the text part of the email is sent to the LSTM model separately. Accuracy with a range between 92.64\% to 98.48\% is achieved by this method.

\section{Background}

This section discusses the details behind the text representation, deep learning architectures and the concept of cost-sensitive model.

\subsection{Text representation}

\subsubsection{Keras Embedding:}
Word embedding takes sequence and similarities into account to convert words into dense vectors of real numbers. Keras provides an embedding layer with few parameters such as dictionary size, embedding size, length of the input sequence and so on. These parameters are hyperparameters and can have an impact on the performance. The weights are taken randomly at first. These weights are tuned during backpropagation with respect to other deep learning layers. Generally, Keras embedding learns embedding of all the words or characters in the training set but the input word or character should be represented by a unique integer. Keras\footnote{https://keras.io/} is neural network library available for the public which has different neural network building blocks like RNN, CNN, etc and also other common layers like dropout, pooling, etc. 

\subsubsection{N-gram:}
From a given sequence of text, the continuous sequences of $N$ items are called are N-gram. N-gram with $N=1$ is known as a unigram and it takes one word/character at once. $N=2$ and $N=3$ are called bigram and trigram respectively and will take two and three words/characters at a time. If $n$ words/characters are to be taken at once then $N$ will be equal to $n$.

\subsection{Machine Learning}

\subsubsection{Naive Bayes:}
is a simple but surprisingly powerful algorithm which is based on Bayes theorem principle. Given the prior information of conditions that may be related with the occasion, it finds the probability of occurrence of an event. 

\subsubsection{Decision Tree:} is another supervised machine learning algorithm. The decision tree is constructed by continuously splitting the data-dependent on certain parameters. Decision trees consist of leaves and nodes where leaves are the results of each decision made and nodes are the decision processes. Iterative Dichotomiser 3 (ID3) algorithm is the most commonly used algorithm to produce these trees. Using Decision Trees both classification and regression are possible for discrete and continuous data.

\subsubsection{AdaBoost:} is an ensemble machine learning classifier like  random forest which utilizes a number of weak classifiers to make a strong classifier. Many machine learning algorithms performance can be boosted using AdaBoost. Training set which is used to iteratively retrain the algorithm is chosen based on the accuracy of previous training. At every iteration, there is a weight given to every trained classifier which is dependent on the accuracy achieved by the classifier. The items that were not correctly classified are given higher weights which makes them have a higher probability in next classifier. Classifier which has an accuracy of 50 percent or more are given zero weight whereas negative weights are given to classifier which has accuracy less the 50 percent. As the number of iterations is increased, the accuracy of the classifier is improved.

\subsubsection{Random Forest (RF):} At first, random forest delivers multiple decision trees. These different decision trees are then merged to get the correct classification. The accuracy of this algorithm is directly proportional to the number of decision trees. Without hyperparameter tuning, random forest gives a very good performance. To reduce overfitting, it utilizes the ensemble learning method while making the decision trees. Different types of data such as binary, numerical or categorical can be given as input to this algorithm.

\subsubsection{Support Vector Machine (SVM):} is another supervised machine learning algorithm which creates a hyperplane to split the data attributes between at least two classes. Every data attribute is projected onto an n-dimensional space. The hyperplane is created in a way such that the distance between the most nearby point of each class and the hyperplane is maximized. Hard margin SVM and soft margin SVM are the two type of SVM where hard margin SVM is the SVM which draws a hyperplane in linear manner whereas soft margin SVM is the SVM which draws the hyperplane in non-linear manner.

\subsection{Deep Learning Architectures}
\subsubsection{Deep Neural Network (DNN):} is an advanced model of classical feed-forward network (FNN). As the name indicates the DNN contains many hidden layers along with the input and output layer. When the number of layer increases in FFN causes the vanishing and exploding gradient issue. To handle these issues, the $ReLU$ non-linear activation was introduced. $ReLU$ helps to protect weights from vanishing by the gradient error. Compared to other non-linear functions, $ReLU$ is more robust to the first-order derivative function since it does not become zero for high positive as well as high negative values.

\subsubsection{Convolutional Neural Network (CNN):} is the most commonly used deep learning architecture in computer vision applications as it has the capability to extract spatial features. The three layers in CNN are convolution layer, pooling layer, and fully connected layer. Convolution layer contains filters that slide over the data to capture the optimal features and these features collectively are termed as a feature map. The dimension of the feature map is high and to reduce the dimension pooling layer is used. Min, max or average are the three pooling operations. Finally, the pooling features are passed into a fully connected layer for classification. For binary classification, $sigmoid$ activation function is used whereas for multiclass classification, $softmax$ activation function is used.

\subsubsection{Long short-term memory (LSTM):} is a special type of recurrent neural network. It takes care of the issue of exploding and vanishing gradient. A cell of LSTM comprises of four major parts namely, input, state cell, three gates and output. The concatenation of the previous output and present input is the input of this LSTM cell. The focal piece of the LSTM cell is called as a state cell which holds the information about the previous sequences. The three gates in an LSTM cell are forget gate, input gate, and output gate. Which information to be remembered or which information to forget is decided by the forget gate. The information relevant to the present input is taken to the cell state by the input gate. What information should be passed as the output of the LSTM cell is decided by the output gate. The output of the present cell gets concatenated with the input of the next cell.

\subsubsection{CNN - LSTM:}

CNN - LSTM architecture was developed for spatial time series prediction problems as LSTM alone cannot handle inputs with spatial structure like images. It consists of CNN layers to exact the features of the input data and LSTM for supporting sequence predictions. In the end, it is connected to a fully connected layer to get the classified output.

\subsection{Cost-sensitive Model}

Models normally treat all samples equally which makes them sensitive to the class imbalance problem. Class imbalance problem arises when there are classes which have very small samples in comparison to other classes in the training data. This problem can be handled using cost-sensitive models. The cost-sensitive deep learning architectures consider all the samples equally. These models give importance to the classes that have more number of samples during training and limits the learning capability to the classes that have very less number of samples. Cost-sensitive learning served as an important method in real-world data mining applications and provides an approach to carefully handle the class imbalance problem. Let's assume that the samples have equal cost at first. $C[i,i]$ indicates the misclassification cost of the class $i$, which is generated using the class distribution as

\begin{equation}
            C[i,i]=\left ( \frac{1}{n_{i}} \right )^{\gamma}
            \end{equation}

Where $\gamma \in [0,1]$. 0 indicates that cost-sensitive deep learning architectures are diminished to cost-insensitive and 1 indicates that $C[i,i]$ is inversely proportional to the class size $n_{j}$.

\section{Description of the Data set}

In this work, three different data sets were utilized for the three use cases. For DGA, domain names have to be classified as legitimate or malicious. For Email, classification result should be either legitimate or spam. For URL, URL should be classified as legitimate or malicious. The data set is divided into train data and test data. The train data set was used to train the models whereas the test data set was used to test the trained models. The train and test dataset of DGA composed of 38,276 legitimate, 53,052 malicious and 12,753 legitimate, 17,690 malicious domain name samples respectively. The legitimate domain names are collected from Alexa\footnote{https://support.alexa.com/hc/en-us/articles/200449834-Does-Alexa-have-a-list-of-its-top-ranked-websites} and OpenDNS\footnote{https://www.opendns.com/} and DGA generated domain names are collected from OSINT Feeds\footnote{https://osint.bambenekconsulting.com/feeds/}. The train dataset is collected from November, 2017 to December 2017 and the test dataset is collected from January 2018 to Feburary 2018. The train and test dataset of email composed of 19,337 legitimate, 24,665 spam and 8,153 legitimate, 10,706 email samples respectively. The train and test dataset of email are collected from Enron\footnote{https://www.cs.cmu.edu/~enron/} and PU\footnote{http://www.aueb.gr/users/ion/data/PU123ACorpora.tar.gz}. The train and test dataset of URL composed of 23,374 legitimate, 11,116 malicious and 1,142 legitimate, 578 malicious samples respectively. 
The legitimate URLs are collected from Alexa.com and DMOZ directory\footnote{https://dmoz-odp.org/} and malicious URLs are collected from  malwareurl.com, Phishtank.com, OpenPhish.org, malwaredomainlist.com, and malwaredomains.com. The train dataset is collected from March, 2018 to April 2018 and the test dataset is collected from September, 2018 to October, 2018. All the datasets are unique as well as the train and test datasets are disjoint to each other.

\section{Proposed Architecture}

\begin{figure}[!htbp]
  \centering
    \includegraphics[width=10cm,height=4cm]{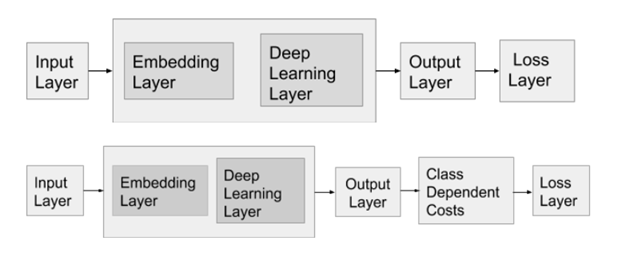}
    \caption{Cost-insensitive and cost-sensitive deep learning based architectures.}
    \label{Fig:1}
\end{figure}

The proposed architecture is shown in Figure \ref{Fig:1}. The diagram located in top shows the training process involved in cost-insensitive deep learning model. The other diagram shows the training process involved in the cost-sensitive deep learning model.
In the cost-sensitive deep learning model, we introduce cost-weights to make the classifier to give importance to the classes which have very less number of samples and give less importance to the classes which have more number of samples. This enables to avoid imbalanced problems in classification.

\section{Experiments, Results, and Observations}

All the classical machine learning algorithms are implemented using Scikit-learn\footnote{https://scikit-learn.org/} and the deep learning models are implemented using TensorFlow\footnote{https://www.tensorflow.org/} with Keras\footnote{https://keras.io/} framework. All the models are trained on GPU enabled TensorFlow. Various statistical measures are utilized in order to evaluate the performance of the proposed classical machine learning and deep learning models.  

We have trained various classical machine learning and deep learning model which can be cost-sensitive or cost-insensitive using the trained datasets. The performance of the trained models is evaluated on test data.

\begin{table}[!htpb]
\centering
\caption{Results for DGA Analysis.}
\label{Tab:2}
\scalebox{0.85}{
\begin{tabular}{|l|l|l|l|l|l|l|l|l|}
\hline
\textbf{Model}                  & \textbf{Accuracy} & \textbf{Precision} & \textbf{Recall} & \textbf{F1-score} & \textbf{TN} & \textbf{FP} & \textbf{FN} & \textbf{TP} \\\hline
\textbf{Naive Bayes}       & 68.1    & 99.3      & 45.5   & 64.4     & 12,700 & 53   & 9,653 & 8037  \\ \hline
\textbf{Decision Tree}  & 79.7    & 76.5      & 93.8   & 84.3     & 7,654  & 5,099 & 1091 & 16,599 \\ \hline
\textbf{AdaBoost}     & 82.8    & 79.2      & 95.6   & 86.6     & 8,300  & 4,453 & 770  & 16,920 \\ \hline
\textbf{RF}  & 84.1    & 80.5      & 95.8   & 87.5     & 8,651  & 4,102 & 736  & 16,954 \\ \hline
\textbf{SVM}     & 85.2    & 81.7      & 96.2   & 88.3     & 8,932  & 3,821 & 680  & 17,010 \\ \hline
\textbf{DNN}          & 86.8    & 83.7      & 96.1   & 89.5     & 9,434  & 3,319 & 688  & 17,002 \\ \hline
\textbf{CNN}           & 94.3    & 92.1      & 98.7   & 95.3     & 11,263 & 1,490 & 233  & 17,457 \\ \hline
\textbf{LSTM}          & 94.4    & 93.0        & 97.6   & 95.3     & 11,457 & 1,296 & 421  & 17,269 \\ \hline
\textbf{CNN-LSTM}      & 95.2    & 93.2      & 99.0     & 96.0       & 11,478 & 1,275 & 174  & 15,716 \\ \hline
\multicolumn{9}{|c|}{\textbf{Cost-sensitive models}}                              \\ \hline
\textbf{CNN}           & 95.4    & 93.2      & 99.5   & 96.2     & 11,464 & 1,289 & 97   & 17,593 \\ \hline
\textbf{LSTM}          & 95.5    & 93.2      & 99.6   & 96.3     & 11,470 & 1,283 & 74   & 17,616 \\ \hline
\textbf{CNN-LSTM  }    & 95.6    & 93.2      & 99.7   & 96.3     & 11,467 & 1,286 & 59   & 17,631 \\ \hline
\end{tabular}}
\end{table}

\begin{table}[!htbp]
\centering
\caption{Results for Email Analysis.}
\label{Tab:3}
\scalebox{0.85}{
\begin{tabular}{|l|l|l|l|l|l|l|l|l|}
\hline
\textbf{Model}                  & \textbf{Accuracy} & \textbf{Precision} & \textbf{Recall} & \textbf{F1-score} & \textbf{TN} & \textbf{FP} & \textbf{FN} & \textbf{TP} \\\hline
\textbf{Naive Bayes}    & 68.8     & 99.4      & 45.3   & 62.2     & 8,122 & 31   & 5,855 & 4,851  \\ \hline
\textbf{Decision Tree}  & 82.9     & 80.0       & 95.4   & 87.1     & 4,521 & 2,527 & 487  & 10,138 \\ \hline
\textbf{AdaBoost}     & 91.3     & 88.6      & 97.1   & 92.7     & 6,815 & 1,338 & 310  & 10,396 \\ \hline
\textbf{RF} & 92.0      & 89.9      & 96.7   & 93.2     & 6,984 & 1,169 & 349  & 10,357 \\ \hline
\textbf{SVM}           & 92.3     & 92.3      & 94.4   & 93.3     & 7,304 & 849  & 599  & 10,107 \\ \hline
\textbf{DNN} & 93.0       & 90.0        & 98.6   & 94.1     & 6,980 & 1,173 & 145  & 10,561 \\ \hline
\textbf{CNN    }       & 93.6     & 92.6      & 96.4   & 94.5     & 7,326 & 827  & 382  & 10,324 \\ \hline
\textbf{LSTM }         & 93.7     & 91.9      & 97.5   & 94.6     & 7,239 & 914  & 270  & 10,436 \\ \hline
\textbf{CNN-LSTM }     & 94.0      & 92.2      & 97.6   & 94.8     & 7,270 & 883  & 253  & 10,453 \\ \hline
\multicolumn{9}{|c|}{\textbf{Cost-sensitive models}}             \\ \hline
\textbf{CNN }          & 94.2     & 92.7      & 97.4   & 95.0      & 7,334 & 819  & 276  & 10,430 \\ \hline
\textbf{LSTM   }       & 94.3     & 92.7      & 97.6   & 95.1     & 7,333 & 820  & 254  & 10,452 \\ \hline
\textbf{CNN-LSTM }     & 94.7     & 92.8      & 98.3   & 95.5     & 7,341 & 812  & 187  & 10,519 \\ \hline
\end{tabular}}
\end{table}

All the models are parameterized as optimal parameters play a significant role in obtaining better performance. For machine learning algorithms, we have not done any hyperparameter tuning. We have used the default parameters of Scikit-learn. To convert text into numerical values, Keras embedding was used with an embedding dimension of 128. Different architectures namely, DNN, CNN, LSTM, and CNN-LSTM are used. For DNN, 4 hidden layers with units 512, 384, 256, 128 and a finally dense layer with 1 hidden unit. In between the hidden layer dropout of 0.01 and batch normalization are used. Dropout was used to reduce overfitting and batch normalization was used to increase the speed. In CNN, 64 filters with filter length 3 and followed by maxpooling with pooling length 2 are used. Followed by a dense layer with 128 hidden units and dropout of 0.3. Finally a dense layer with one hidden unit. LSTM contains 128 memory blocks followed by dropout of 0.3 and finally dense layer with one hidden unit. In CNN-LSTM architecture, we connected CNN network with LSTM network. CNN has 64 filters with filter length 3, followed by a maxpooling layer having pooling length 2. Followed by LSTM network having 50 memory blocks and finally a dense layer with one hidden unit is added. All the experiments are run till 100 epochs with a learning rate of 0.01, and $adam$ optimizer.

\begin{table}[!htbp]
\centering
\caption{Results for URL Analysis.}
\label{Tab:4}
\scalebox{0.85}{\begin{tabular}{|l|l|l|l|l|l|l|l|l|}
\hline
\textbf{Model}                  & \textbf{Accuracy} & \textbf{Precision} & \textbf{Recall} & \textbf{F1-score} & \textbf{TN} & \textbf{FP} & \textbf{FN} & \textbf{TP} \\\hline
\textbf{Naive Bayes}            &  45.1        & 37.9          & 98.8      &  54.7        &  205  &  937  & 7   &  571  \\\hline

\textbf{Decision Tree}          & 81.8       &   73.3        &  72.1      &  72.7        &  990  &  152  & 161   & 417   \\\hline
\textbf{AdaBoost}             &    87.1      &  83.8         & 76.3       & 79.9         & 1,057   &  85  &  137  &  441  \\\hline
\textbf{RF}         &   90.0       &   90.6        &  78.4      &      84.0    & 1,095    & 47    &  125  &   453 \\\hline
\textbf{SVM} &   81.0       &  88.4         &  50.0      &    63.9      & 1,104   & 38   &  289  &  289  \\\hline
\textbf{DNN}   &  90.8        &   92.5        &    79.1    & 85.3         &  1,105  &  37  &  121  & 457   \\\hline
\textbf{CNN}                  &   92.9       &   91.9        &    86.5    &    89.1      &  1,098  &   44 & 78   &  500  \\\hline
\textbf{LSTM}                &   93.4       &     97.2      &   82.7     &    89.3      &  1,128  & 14   & 100   &  478  \\\hline
\textbf{CNN-LSTM}               &   94.4       &   96.5        & 86.5        &  91.2         &  1,124  &   18 &    78 &  500   \\\hline
\multicolumn{9}{|c|}{\textbf{Cost-sensitive models}}                                           \\\hline
\textbf{CNN}                    &   93.4       &  96.0         &  83.7      &  89.5        &  1,122  &  20  & 94   &   484 \\\hline
\textbf{LSTM}                  &      94.5    &    93.7       & 89.8        &   91.7       &  1,107 &  35  &   59 &  519  \\\hline
\textbf{CNN-LSTM}               & 94.7         & 93.1        &   91.0     &     92.0     & 1,103   & 39   & 52   & 526  \\\hline
\end{tabular}}
\end{table}

The detailed performance analysis of all the models are reported in Table \ref{Tab:2} for DGA analysis, Table \ref{Tab:3} for Email analysis, and Table \ref{Tab:4} for URL analysis. As shown in the tables, the performance of deep learning models is better than the machine learning models. More importantly, the performance of cost-sensitive deep learning models is better than the cost-insensitive models. This is primarily because cost-sensitive models can give certain weights to the classes which helps to reduced overfitting and underfitting during training. We can see in all the three Tables that the cost-sensitive hybrid network of CNN-LSTM performed better than the other network like CNN and LSTM.

\section{Conclusion and Future Work}

This paper proposes a generalized cost-sensitive deep learning model for  Cyber Security use cases such as DGA, Email, and URL. However, the model can be applied on other  Cyber Security use cases also. The cost-sensitive hybrid model composed of CNN and LSTM can extract spatial and temporal features and can obtain better perform on any type of data sets. Implementing this model in real time data analysis with Big data and Streaming can be considered a good direction for future work.

\section*{Acknowledgements}

This research was supported in part by Paramount Computer Systems and Lakhshya Cyber Security Labs. We are grateful to NVIDIA India, for the GPU hardware support to research grant. We are also grateful to Computational Engineering and Networking (CEN) department for encouraging the research.

\end{document}